%% file: LecomteQuatrini-wollic09.tex
\documentclass[oribibl, english]{llncs}
\usepackage{geometry}
\usepackage{amsfonts,amsmath,latexsym,amssymb,amstext}
\usepackage{babel}
\usepackage{amssymb, eufrak}
\usepackage{times}
\usepackage{bussproofs}
\input{nprooftree.tex}

\newcommand\seq\vdash 
   
\newcommand\tms\otimes
\newcommand\ra\rightarrow
\newcommand\Ra\Rightarrow
\newcommand\llts\otimes
\newcommand\pt\bullet 
\newcommand\lts\bullet

\newcommand{\linefle}{\mathrel{-\hspace{-0.2em} \circ }}

\newcommand\dw\downarrow

\newtheorem{rem}{Remark}
\begin{document}
\input{qobitree}
\title{Ludics and Its Applications to Natural Language Semantics}
\author{Alain Lecomte\inst{1}\and Myriam Quatrini\inst{2}}
\institute{UMR "Structures Formelles de la Langue", CNRS-Universit\'e Paris 8 - Vincennes-Saint-Denis
\and UMR "Institut de Math\'ematiques de Luminy", CNRS-Aix-Marseille Universit\' e}

\date{}
\maketitle
\begin{abstract}
\noindent
Proofs in Ludics, have an interpretation provided by their  {\it counter-proofs}, that is the objects they interact with. We shall follow the same idea by proposing that sentence meanings are given by the {\it counter-meanings} they are opposed to in a dialectical interaction. In this aim, we shall develop many concepts of Ludics like \emph{designs} (which generalize proofs), \emph{cut-nets}, \emph{orthogonality} and \emph{behaviours} (that is sets of designs which are equal to their bi-orthogonal). Behaviours give statements their interactive meaning. Such a conception may be viewed at the intersection between \emph{proof-theoretic} and \emph{game-theoretical} accounts of semantics, but it enlarges them by allowing to deal with possibly infinite processes instead of getting stuck to an « atomic » level when decomposing a formula.
\end{abstract}
\section{Meanings, Proofs and Games}
\noindent
The dominant trend in Natural Language Semantics is based on Frege's conceptions on Logics and Language according to which the meaning of a sentence may be expressed in terms of its truth conditions.  There is however an alternative conception  according to which we don't find meanings in truth conditions but in \emph{proofs}, particularly expressed by the \emph{Brouwer-Heyting-Kolmogorov} (BHK) - interpretation. This conception has been used in philosophy, linguistics and mathematics. In Natural Language Semantics, it has been for instance developed by Martin-L{\"o}f, Sundholm and Ranta (\cite{ML84, Sund86, Ran94}), but this framework  is limited because proofs are finite objects. At a certain stage of the proof of a formula, atomic formulae are obtained, but \emph{what is the proof of an atomic formula}? 
%Ranta argues that \emph{a proof object is (simply) an object that makes a proposition true}, and that the sentence \emph{there is a railway from Moscow to Hong Kong} has as its "proofs"...  railways from Moscow to Hong Kong, but we don't find such "solution" satisfactory. In fact 
Actually, we expect a proof of a sentence to be an object of the same symbolic nature as the sentence. There is no way to escape from language or from mind to directly reach the external world.\\
Other similar attempts to provide a foundation for meaning in natural language are based on works by Hintikka, Kulas and Sandu (\cite{HinKul83, Hintikka}). In their interpretation, meanings are provided by strategies in a \emph{language game}. Their views meet Wittgenstein's according to which \emph{meaning is use} and the use of language is showed in language games. But still those accounts meet difficulties when dealing with atomic sentences: in this case, the logician is obliged to refer to some model (in the traditional sense of Model Theory) in order to evaluate the truth value of an atomic sentence.\\
Still more seriously, they only take in consideration games which are of a very particular kind: they are oriented towards the notions of \emph{winner}, \emph{winning strategy}, \emph{score} and \emph{pay-off} function, contrarily to what Wittgenstein suggested in his \emph{Philosophische Untersuchungen} when he spoke of games for a very large family (even mere pastimes). Neither Wittgenstein's games do refer to \emph{a priori} rules which would be attached, like in GTS, with logical particles (cf. \cite{Piet07}).\\ 
Finally, none of these alternatives to the truth-conditions based framework ever envisaged to take proofs or games as infinite devices (or partial and underspecified ones) and of course none of these traditions took into account the fact that proofs and strategies can be the \emph{same} objects, simply viewed from different angles.\\
If this concerns formalized theories of meaning, what to say of theories of meaning which have not been formalized, like that of \emph{argumentative} meaning, in O. Ducrot's sense (\cite{Ducrot}). Ducrot pointed out the so-called \emph{polyphonic} aspect of language, that is the fact that utterrances are not simple statements which are confronted with "reality", but dynamical processes which are oriented towards possible or impossible continuations (for instance \emph{I have a few books} cannot be pursued by $^*$\emph{and even none}, while \emph{He read few books} may be). In the same way, \emph{dialogues} may be studied according to what utterrance may be an appropriate reply to another one, and what may not be.\\
In this paper, we shall present some applications of Ludics to these topics. 
In a nutshell, proofs in Ludics, have an interpretation provided by their  counter-proofs, that is the objects they interact with. We shall follow the same idea by proposing that sentence meanings are given by the {\it counter-meanings} they are opposed to in a dialectical interaction.
%In this paper we describe  a simplified presentation of Ludics as a hypersequentialized sequent calculus.  Then, we present an account of {\it Logical Forms} following directions suggested by Ludics~: the meaning of a sentence is given by all the logical forms which correctly interact with its own logical form.  
%Precisely we define meanings as sets of logical forms stable by {\it bi-orthogonality} or as: \emph{behaviours} (according to the Ludics terminology). 
%{\it Préciser pourquoi la Ludique nous paraît être un cadre pertinent : parce que c'est une théorie de l'interaction et nous posons que l'ancrage dialogique est essentiel pour approcher les phénomènes langagiers ; parce qu'elle nous permet de nous libérer en partie de la notion de proposition ou formule au profit de celle de proto-formule (pour parler d'un niveau de formulation logiquement décomposable), de la notion de preuves au profit de la notion de proto-preuves (ou processus ou de support d'interaction)}.
\section{Dialogues and Ludics}
\subsection{Ludics : a Theory of Interaction}
Ludics can be sum up as an  interaction theory, formulated by J.-Y. Girard (\cite {locus}) as the issue of several changes of paradigms in Proof Theory : from   provability to   computation, then from computation to interaction. The first change of paradigm arose with the intuitionnistic logic, while the second was due to the development  of linear logic.\\
Starting from a geometrical viewpoint on proofs, which provided an internal approach to the dynamics of proofs, Ludics takes the notion of  interaction (that is the cut rule and its process of elimination) as \emph{primitive}. Therefore, it simply starts from {\it loci}, or adresses (\emph{where} interaction can take place) and \emph{formulae} are given up, at least for a while, since the challenge is to regain them at the output of the construction. \emph{Proto-formulae} are used as mere scaffoldings for building the main objects we shall deal with (the designs).\\
\noindent{\bf The central object of Ludics: the design}
Using the metaphor of Games,  a design can be understood as a {\it strategy}, i.e. as a set of {\it plays} (or {\it chronicles}) ending by answers of Proponent against the  moves planned by  Opponent.  The plays are  alternated sequences of {\it moves} ({\it actions}). A move is defined as a 3-uple consisting in
\begin{itemize}
\item a polarity (positive for Proponent, negative for Opponent),
\item an adress or  {\it locus}, coded by a finite sequence of integers (denoted by $\xi$, $\rho$, $\sigma$ \dots),  where the move is said to be \emph{anchored},
\item a finite set of  integers, or {\it ramification} which indicates the positions which can be reached in one step. A unusual positive move is also possible~: the {\it da\" \i mon}, which may end up a play.
\end{itemize}
Positions are organized in  {\it forks}, which are presented under the general form: 
 $\Gamma\vdash\Delta$%. Roughtly speaking, it is an organisation of positions from which the play can start~
; where $\Gamma$ and $\Delta$ are finite sets of loci such that $\Gamma$ is either the empty set or a singleton. The fork corresponding to the starting position is called the {\it base} of the design.   When $\Gamma$ is not empty, the following (opponent) move starts from the only element it contains, and the fork is said to be \emph{negative},  in the other case, Proponent chooses the locus in $\Delta$ from where it starts and the fork is said to be \emph{positive}.
\vspace{0,2em}

Perhaps this may seem not new with regards to \emph{GTS}, let us notice however that moves are defined abstractly, independently from any particular connective or quantifier, and that at each step, the whole history of the previous moves is available.

Now, more importantly, a design may be also seen as  a \emph{proof search}  in some linear formal system, according to the following methodological choices:
\begin{itemize}
\item the object we are building not only provides a proof, but at the same time, contributes to the determination of the formula which is proved. Loci point out the position where such a formula could be located, and at the same time, the "logical" decomposition this formula could have,
\item by means of the focalisation property discovered by Andr\'eoli \cite{And92} according to which in linear logic, it is always possible to draw a proof by following a strict discipline (focusing) which amounts to grouping together successive blocks of rule applications of the same polarity, it is possible to have only two rules (one positive and one negative).
\item it may happen that the research be not successful. In this case, one may give up the proof search, thus using a specific non logical rule (or \emph{paralogism}) : the \emph{da{\"\i}mon} rule.
\end{itemize}
A design can therefore be represented by a tree of forks, built by means of three rules~:

\noindent
 {\bf Da\"\i mon}
$$
\displaylines{
\prooftree
\justifies
\vdash \Lambda
\using
\dagger
\endprooftree
}
$$
\noindent
{\bf Negative rule}
$$
\displaylines{
\prooftree
... \quad\vdash \xi\star J, \Lambda_J \quad...
\justifies
\xi \vdash  \Lambda
\using
(-, \xi, \cal{N})
\endprooftree
}
$$
{\bf Positive rule}
$$
\displaylines{
\prooftree
... \quad\xi\star i \vdash \Lambda_i \quad...
\justifies
\vdash \xi, \Lambda
\using
(+, \xi, I)
\endprooftree
}
$$
where $I$ and $J$ are finite subsets of ${\mathbb N}$, $i\in I$, with the $\Lambda_i$ pairwise disjoints, $\cal{N}$ is a set ({\it possibly infinite}) of finite subsets of ${\mathbb N}$, each $J$ of the negative rule being an element of this set, all the $\Lambda_J$  are included in $\Lambda$, and moreover each base sequent is well formed in the sense that all addresses are pairwise disjoint.\\

\noindent{\bf The Fax} Since we have not yet introduced formulae,  there is no opportunity to use \emph{axiom}-links. Instead, we will have a particular design based on a fork $\xi\vdash\xi'$. Roughly speaking, this design ensures that both loci $\xi$ and $\xi'$ could be locations of a same formula. That means that as soon as a logical decomposition  may be handled on the right hand side, the same may also be handled on the left hand side. Such a design, 
called ${\cal F}ax$,  is recursively defined  as follows:
$$
\displaylines{
Fax_{\xi, \xi'} =
\prooftree
...
\[
...
\[Fax_{\xi'_{i}, \xi_{i}}
\justifies
\xi'\star i\vdash \xi\star i
\]
...
\justifies
\vdash \xi\star J, \xi' 
\using
(+, \xi', J)\]
...
\justifies
\xi \vdash \xi'
\using
(-, \xi, {\cal P}_f({\mathbb N}))
\endprooftree
}
$$
At the first (negative) step, the negative \emph{locus} is distributed over all the finite subsets of ${\mathbb N}$, then for each set of addresses (relative to some $J$), the positive locus $\xi'$ is chosen and gives rise to a subaddress $\xi'\star i$ for each $i\in J_k$, and the same machinery is relaunched for the new loci obtained.\\

\noindent{\bf Defining interaction}
Interaction  is concretely expressed by a coincidence of two loci in dual position in the bases of two designs. This creates a dynamics of rewriting of the cut-net of the two designs, called, as usual, {\it normalisation}. We sum up  this process as follows: the cut link is duplicated and propagates over all immediate \emph{subloci} of the initial cut \emph{locus} as long as the action anchored on the positive fork  containing the cut-locus corresponds to one of the actions anchored on the negative one. The process terminates either when the positive action anchored on the positive cut-fork is the \emph{da{\"\i}mon}, in which case we obtain a design with the same base as the starting cut-net, or when it happens that in fact, no negative action corresponds to the positive one. In the later case, the process fails (or \emph{diverges}). The process may not terminate since designs are not necessarily  finite objects.\\
 When the normalization between two designs ${\cal D}$ and ${\cal E}$ (respectively based on $\vdash
\xi$ and $\xi\vdash$) succeeds, the designs  are said to be {\it {orthogonal}}, and we note: ${\cal D}~\bot~{\cal E}$.
 In this case, normalization ends up on the particular design : 
$$
\prooftree 
\justifies 
\quad\vdash\quad
\using 
[\dag] 
\endprooftree
$$
Let ${\cal D}$ be a design, ${\cal D}^\bot$ denotes the set of all its orthogonal designs. 
It is then possible to compare two designs according to their counter-designs. We set ${\cal D}\prec{\cal E}$ when ${\cal D}^\perp\subset{\cal E}^\perp$.\\

\noindent
 \emph{The separation theorem \cite{locus} ensures that this relation of preorder is an order, so that a design is exactly defined by its orthogonal}.\\

\noindent{\bf Behaviours}
One of the main virtues of this "deconstruction" is to help us rebuilding Logic. 
\begin{itemize}
\item Formulas are now some sets of designs. They are exactly those which are closed (or stable) by interaction, that is those which are equal to their \emph{bi-orthogonal}. Technically, they are called \emph{behaviours}.
\item The usual  connectives of Linear Logic are then recoverable, with the very nice property of \emph{internal completeness}. That is : the bi-closure is useless for all linear connectives. For example, every design in a behaviour ${\bf C}\oplus {\bf D}$ may be obtained by taking either a design in ${\bf C}$ or a design in ${\bf D}$. 
\item Finally, \emph{proofs} will be now designs satisfiying some properties, in particular  that of not using the da{\"\i}mon rule.
\end{itemize}
\subsection{Ludics as a Formal Framework for Dialogues}
Concerning dialogues, let us focalize on the mere \emph{supports} of the interaction. That is the \emph{locus} where a speech turn is anchored (among the \emph{loci} previously created) and the \emph{loci} that it creates, which are also those which may be used later on. \\
Because Ludics may display the history of the dialogue by means of \emph{chronicles}, and it takes into account the strategies of any speaker by means of \emph{designs}, it allows us to see a 
dialogue as the result of an interaction between the strategies of two speakers. In that case, the rules have the following interpretation:
\begin{itemize}
\item when being \emph{active} (that is using a positive rule), a speaker chooses a \emph{locus} and therefore has an active role,
\item when being \emph{negative} (that is using a negative rule), s/he has no choice and has a passive role
\end{itemize}
If, therefore, positive steps are understood as moves where the intervener asks a question or makes an assertion, and negative steps as moves where s/he is apparently passive, recording an assertion and planning a further reply, positive actions of one speaker are not opposed to positive actions of the other one (as it is the case in most formal accounts of dialogue, even the logical ones) but to negative ones of the other. 
This point meets an important requirement formulated by Ducrot according to whom ''the semantic value of an utterrance is built by allusion to the possibility of another utterrance (the utterrance of the Other speaker)''.\\

\noindent{\bf Examples}
\begin{itemize}
\item The following example is deliberately simple, and only given for a pedagogic purpose.

\noindent
Let us consider the following dialogue between Annie and Barbara:

 { {\bf A } : did you meet some friends yesterday evening to the party ?
  {\bf B } : I only saw Bruno and Pierre.
   {\bf A } : Was Pierre still as nice as during the last year ? 
  {\bf B } :  Yes, he did. 
  {\bf A } :  That is what I wanted to know. 
   }
%{\bf Idea}:  to sketch {\underline {not the dialogue in itself}} but  its support (as a set of locations). \\
%Which {\it loci} are used where are anchored the successive statements, which ones are created?

\noindent Such an exchange is represented by an interaction between two designs : one is seen from the point of view of {\bf A} and the other from the point of view of {\bf B}:
 \begin{center}
 \begin{tabular}{ccc}
From A's point of view  &\hspace{5em} & From B's point of view \\ 
 $\hrulefill_\dag$&\hspace{5em}& \\
$ \vdash0.1.1.1.1$ & \hspace{5em} &$0.1.1.1.1\vdash$\hspace{3em} \\
 $0.1.1.1\vdash $  &\hspace{5em} & $\vdash 0.1.1.1$\hspace{3em} \\
$\vdash 0.1.1, 0.1.2$  &\hspace{5em} &$0.1.1\vdash$ \hspace{2em}$0.1.2\vdash$\\
$0.1\vdash$ & \hspace{5em} &$\vdash0.1$\\ 
 $ \vdash 0$&\hspace{5em} &$0\vdash $\\
 \end{tabular}
 \end{center}
 
The trace of the interaction (the cut between the two loci $0$)  is the alternated sequence of actions:
$
(+,0,\{1\})(-,0.1,\{1,2\})(+,0.1.1,\{1\})(-,0.1.1.1,\{1\})\dag
$.

In this case the normalisation ends up on the da{\"\i}mon. The interaction converges.\\

\item The second example is taken from Schopenhauer's ''Dialectica eristica'' (or ''The Art of Always Being Right'') which provides a series of so-called {\it stratagems} in order to be always right in a debate. It formalizes the first given stratagem.
 
 {\it ``I asserted that the English were excellent in drama. My opponent attempted to give an instance of the contrary, and replied that it was a well-known fact that in opera, they were bad. I repelled the attack by reminding him that, for me, dramatic art only covered tragedy and  comedy ....''  }

We give an account of this dialogue by the following interaction:

 \begin{center}
 \shortstack{
 \shortstack{ $\vdash\xi.1.1$\hspace{2em}
 $\vdash\xi.1.2\quad$\\
             $\hrulefill_ C$\\
             $\xi.1\vdash$\\
             $\quad\hrulefill_ A\quad$\\
             $\vdash\xi$}
             \hspace{2em}
             \shortstack{$\xi.1.3\vdash$\\
             $\hrulefill_ B$\\
             $\vdash\xi.1$\\
             $\hrulefill$\\
             $\xi\vdash$}\\
             $\qquad\qquad\hrulefill\quad$}
             \end{center}

\noindent{\footnotesize Where the action  A corresponds with the claim: ''The English are excellent in drama'' ; the action B with ``I disagree, it is a well-known fact that in opera, they could do nothing at all.'' and the action C with ``But by  dramatic art, I only mean tragedy and comedy.'' }

   \noindent
Of course, the net built with these two designs does not converge. In fact, things don't happen this way: initially, the set of loci the first speaker has in mind could also cover \emph{opera}. What happens when willing to repel the attack is retracting one branch (or replay the game according to a different strategy). This leads us to enter more deeply into the decomposition of dialogues and in what we consider as {\it units} of action.
\end{itemize}

% \vspace{0,5em}
 
\noindent 
%The formal decomposition of dialogues may be considered, in Ludics, at various levels of  granularity: For an elementary decomposition, a dialogue is seen as an alternation of signed interventions~; the speaker anchors its intervention on a locus (a fixed position) created by the previous interventions. Some figures (trees / designs of Ludics) emerge ; also, the supports of the sequence of interventions appear ; we can locate in it the trace of the actual interaction of the dialogue in progress. We can then refine this approach and decompose the interventions themselves, with regard to the way they are dynamically built. 
While, at the most elementary level, which is relevant as long as the dialogues we consider are  simple (for instance exchanges of information), the interaction is between elementary actions, those elementary actions are replaced by (sub)-designs  as soon as we are concerned by dialogues of a more complex nature like controversies.
%To associate a whole design or a cut-net with the intervention of the speaker is the sign that this exchange could be broken down into more elementary ones.   We shall perform  this operation in order to take care of the dynamics of these exchanges. %The example of the presupposition will illustrate such an extension.  An intervention is no more represented by actions (moves) but by  whole designs (plays). 

\begin{itemize}
\item A third example comes from Aristotle's \emph{Sophistical Refutations}, where it is given the name \emph{multiple questions}.

\noindent
Let us imagine a judge asking a man the question: 

{\it ``Do you still beat your father ?''}. 

\noindent
The judge asks a question that presupposes something that has not necessarily  been accepted by the man. S/he imposes to him the following implicit exchange:

{\it - ``Do you beat your father?''\   - `` Yes''\    - `` Do you stop  beating him~?''}. 

\noindent This exchange between the judge $J$ and the man $D$ must be represented by the following interaction :
\begin{center}
\shortstack{
\shortstack{
\shortstack{${\bf \xi.0.1.0}\vdash$\\
$\hrulefill$\\
$\vdash {\bf \xi.0.1}$}\hspace{1em}$\vdash\xi.0.2$\\
$\hrulefill$\\
$ {\bf\xi.0}\vdash$\\
$\hrulefill$\\
$\vdash {\bf \xi}$}
\hspace{1em}
\shortstack{$\vdash \xi.0.1.0$\\
$\hrulefill$\\
$ \xi.0.1\vdash$ \\
$\hrulefill$\\
$\vdash \xi.0$\\
$\hrulefill$\\
$\xi\vdash$}\\
$\qquad\qquad\hrulefill\qquad$\\
$J$\qquad\qquad\qquad $D$}
\end{center}
 
\noindent In fact, the judge utterance: - {\it ``Do you still beat your father ?''}  contains what we call nowadays a \emph{presupposition}. It  can't therefore be represented by \emph{a single action}, but by \emph{the whole chronicle}: $(+,\xi,\{0\})$ $(-,\xi.0,\{1\})$ $(+,\xi.0.1,\{0\})$. This enables us to give an account of the fact that one of the \emph{loci} where  the interaction might continue is in fact not available ; in some sense the action giving this possibility is skipped, some successive one is immediately proposed and, by this way, constrains the answers. 
\end{itemize}
 %So $J$ forbids addressee a branch who was due to him (the possibility to answer `` No'' ). If D agrees to answer according to this configuration (without diverging) he is trapped: he has to record the whole chronicle\\
%\vspace{0,5em}
\noindent The ludical approach thus allows us to get a formalized conception of stratagems and fallacies, something which appeared out of reach for many researchers (see for instance \cite{Ham70}). Moreover, we claim that it could improve some issues in formal semantics, like we try to show it in the following section.
\section{Logical Forms and Ludics}

\noindent In the sequel, we propose a conception of interactive meaning based on Ludics. In the same way a design is defined by its orthogonal (according to the separation theorem), we may postulate that the meaning of a sentence is given by the set of all its dual sentences: that is all the sentences with which the interaction converges. For this purpose, we associate a behaviour or a family of behaviours with a sentence. Such behaviours are built in a compositional way, like in standard formal semantics, but their ultimate components are neither \emph{atoms} nor \emph{atomic formulae}, like in the Intensional Logic Montague was using. Let us underline the points which are slightly different and new and which could  favourably extend the standard models of semantics:
\begin{itemize}
\item The fact that the mathematical object associated with the meaning of a sentence may be \emph{more and more refined} seems to us very important. Such an objective is realized because of  the order on designs involved by the \emph{separation theorem}, which enables one to explore more and more precisely the argumentative potential of a sentence. Moreover, new designs may always be added to such an object,  thus enlarging our conception of meaning.
\item The fact that Ludics strictly encompasses logic and that logical concepts like formulas, proofs or connectives are defined in a  world which is larger than the strictly logical one (let us remember that we have \emph{paralogisms} like the \emph{da{\"\i}mon}, and counter-proofs in that world!) makes us to expect more freedom in defining ''logical'' forms. For instance it may be the case that behaviours are composed by means of a non-logical operator (but which could nevertheless be interpreted).
\end{itemize}
% We start foregoing a track to deal  with speech act. 

%From the previous section, we may draw conclusions about interaction in language. The logical form of a sentence may be seen as a behaviour. As such, it is a set of designs which interact in the same way with regards to other designs, and a design is itself an abstraction of \emph{paraproofs} that, from now on, we  could term more appropriately \emph{parameanings}. Parameanings may interact in order to end up with the \emph{daimon} (in which case they are orthogonal) but they may also interact in order to produce side-effects which give 'answers' as residuals. Of course sometimes the interaction may diverge. That is to say that interaction (normalization) may be used as providing criteria for distinguishing readings in case of ambiguous sentences.\
%\vspace{0,5em}
\noindent The following example illustrates a classical problem of ambiguïty (\emph{scope} ambiguity).\subsection{Meaning through Dual Sentences}
%\subsection{Logical forms in hypersequentialized logic}
\noindent
The meaning of a sentence is given by all the utterances which correctly interact (that means : \emph{converge}) with it.

\noindent Let us consider the statement (from now on denoted by $S$): 
``Every linguist speaks some african language''.
Usually two logical forms can be associated with such a sentence $S$, depending on whether 
{\it some} has the  narrow or  the wide scope. Namely:

  $S_1=\quad\forall x (L(x)\Rightarrow \exists y (A(y)\wedge P(x,y)))$
  
  $S_2=\quad\exists y (A(y) \wedge\forall x (L(x)\Rightarrow  P(x,y)))$

\noindent
where $L(x)$ means ''x is a linguist'' , $A(y)$ means ''y is an african language'' and $P(x, y)$ means ''x speaks y''. 

\noindent When ''some'' has the narrow scope, we assume that the logical form converges with the LF  of sentences like:

(1)  There is a linguist who does not know any african language.

(2)  Does even John, who is a linguist, speak an african language ?

 (3)  Which is the African language spoken by John ? 
 
\noindent
On the opposite,  if ''some'' has the wide scope, the logical form converges with~:
 
(4)  There is no african language which is spoken by all the linguists.

(5)  Which african language every linguist speaks ? 

\subsection{Meaning as a Set of Justifications}
We materialize the claim according to which  meaning is equated with a set of dual sentences by associating with the meaning of $S$ a set of designs. Such designs may be seen as {\it justifications} of $S$. That is the supports  of the dialogues during which a speaker $P$ asserts and   justifies  the statement $S$  against an adressee $O$ who has several tests at his/her disposal. 
%That is then a design which interacts in a convergent way with counter-designs exploring it.
%What does a design ${\cal D}$ which would be a justification of the statement $S$ look like ? 
%If we follow the interpretation of designs as strategies in a game (one intervention of the Proponent/speaker $P$  is a positive action (seen from its point of view) ; the interventions of its opponent/addressee $O$ is a negative action (from the point of view of the Proponent)) this question may be translated into~: 

\noindent
 %\begin{center}
%% If we follow the representation of the support of dialogue by a design as it was proposed in the previous section, this question may be translated into: 
%How to justify the statement ``every linguist speaks at least one african language'' against an opponent who has several tests at his/her disposal ?

Let us make such a design, based on  the arbritary fork $\vdash 0$, more precise: % \end{center}
\begin{itemize}
\item the  first action corresponds to the assertion of $S$. Its \emph{ramification} is a singleton ; only one locus is created for continuing the interaction. Nevertheless, the speaker who has to anticipate the reactions of his/her adressee is  committed to one of the readings of $S$. S/he is ready to assume one of the two possibilities, the wide or the narrow scope for ''some''. This is taken into account by distinguishing between two possible first actions, that we symbolize for instance by $(+,0,\{0\})$ and $(+,0,\{1\})$. It is then possible to distinguish between two kinds of designs, considered as justifications of $S$ according to the choice of the first action.  

\item let us for instance focus on the first reading of $S$. We then simulate an interaction between $P$ and $O$ who tries to negate $P$'s claim:
 
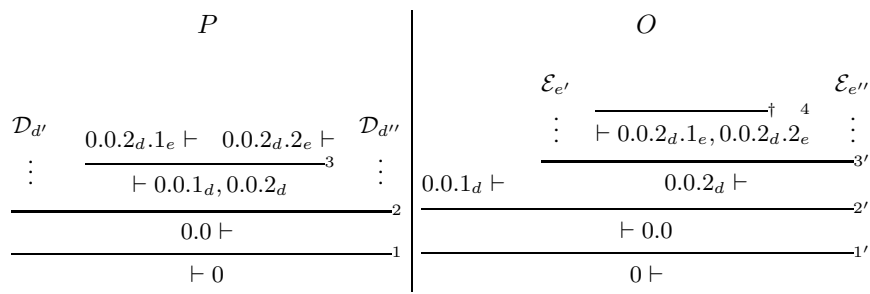
\begin{center}
\begin{tabular}{c|c}
$P$&$O$\\
\\
  \footnotesize{
\shortstack{\shortstack{${\cal D}_{d'}$\\
$\vdots$\\
$\quad$}
\hspace{1em}
\shortstack{$0.0.2_d.1_e\vdash$\hspace{1em}$0.0.2_d.2_e\vdash$\\
$\hrulefill_3$\\
$\vdash  0.0.1_d, 0.0.2_d$}\hspace{1em}\shortstack{${\cal D}_{d''}$\\
$\vdots$\\
$\quad$}
 \\
$\hrulefill_2$\\
$0.0\vdash$\\
$\hrulefill_1$\\
$\vdash 0$}
}
&
\footnotesize{
\shortstack{
\shortstack{ 
$0.0.1_d\vdash$}
\hspace{1em}
\shortstack{\shortstack{${\cal E}_{e'}$\\
$\vdots$}\hspace{1em}\shortstack{$\hrulefill_{\dag\quad4}$\\
$\vdash 0.0.2_d.1_e, 0.0.2_d.2_e$}\hspace{1em}\shortstack{${\cal E}_{e''}$\\
$\vdots$}\\
$\hrulefill_{3'}$\\
            $0.0.2_d\vdash$}\\
             $\hrulefill_{2'}$\\
            $\vdash 0.0$\\
            $\hrulefill_{1'}$\\
            $0\vdash$}
            } \\
            \end{tabular}
            \end{center}
    \vspace{0,5em}
 \end{itemize}
The normalisation stages may be commented as follows:
\begin{enumerate}
\item[1 ] $P$ asserts $S$ and is ready to 
continue the interaction with the first reading of $S$
\item[1'] $O$ records the claim made by $P$  and is ready to answer it.
{\footnotesize Notice that if $O$ had been ready to answer according to the second reading, its action would have been $(-,0,\{1\})$ and the interaction would have diverged}
\item[2 ] $P$ is ready to give justifications for any individual~: $d$,$d'$,\dots
\item[2'] $O$ proposes an individual $d$ (arguing that $d$ is a linguist (localized in $0.0.1_d$) and that $d$ doesn't know any african language (localized in $0.0.2_d$))
\item[3 ] $P$ exhibits some language $e$  (arguing that $e$ is an african language  and $d$ speaks $e$) 
\item[3'] at the same time, $O$ is ready  to receive such a claim by $P$ for some language among  $e'$,$e$, $e''$ \dots
\item[4 ] if $P$ has given some language $e$ such that $d$ speaks it, $O$ may be ready to give up.
\end{enumerate}
\noindent
Thus, the interaction between ''Every linguist speaks some african language'' and the attempt to negate it ''There is some linguist which doesn't speak any african language'' normalizes.

\noindent   
    Let us denote by ${\cal D}$ the foregoing design of $P$. We could also find another design as justification of $S$ with its first reading :  $P$ may ask to {\bf check} if $d$ is really a linguist,  $O$ may ask to {\bf check} if $d$ really speaks $e$ and so on… thus providing a deeper interaction. Further exchanges may enter into debates on what it means for a person to be a linguist, or on what it means for a language to be an african one, or on what it means for a person and a language to be such that the person speaks the language and so on...
    
   \noindent 
In any way, if ${\mathbb S}_1$ denotes the set of designs representing the first reading of $S$ and if ${\mathbb S}_2$ denotes the set of designs representing the second one, the set of designs representing the meaning of $S$ is the union of both sets : ${\mathbb S}={\mathbb S}_1\cup {\mathbb S}_2$.
%This design can be enough to justify $S$  in its first reading and we could set that ${\mathbb S}_1$ is the behaviour generated by ${\cal D}$, namely ${\cal D}^{\perp\perp}$. If we denote by ${\mathbb S}_2$ a behaviour associated with the second reading of $S$, we obtain that the behaviour ${\mathbb S}$ associated with $S$ is equal to ${\mathbb S}_1\cup {\mathbb S}_2$. The fact that this set of designs is a behaviour is justified below.
%This design ${\cal D}$ is in fact \emph{minimal} in the following sense : it is only based on the \emph{logical} articulation of $S_1$. Nevertheless, ${\mathbb S}_1$ contains also some  less defined ( but more informative) designs (enable to interact with more precise questions).\\
\subsection{Meaning  as Behaviour}
\noindent The previous attempt to associate a set of design with the meaning of $S$ is still general and imprecise. 
%But it enables one to retrieve the notion of meaning as {\it logical form}.
${\cal D}$ actually belongs to the following behaviour\footnote{$\forall$ and $\exists$ are used here because of their intuitive appeal, but in fact they stand for the generalized additives  
connectives $\&_x$ and $\oplus_y$ (cf. annex). There is nevertheless a slight difference between both pairs of concepts: strictly speaking, in Ludics the correct use of first order quantifiers with regards to mathematical formulas would involve a uniformity property (\cite{first-order}) which is  
neither relevant nor satisﬁed here.}:
 \begin{center}
$\forall x (\downarrow {\mathbb L}(x)\linefle \exists y(\downarrow {\mathbb A}(y)\otimes \downarrow {\mathbb P}(x,y)))$
\end{center} 
provided that $ {\mathbb L}(x)$ , ${\mathbb A}(y)$ and $ {\mathbb P}(x,y)))$ are behaviours.
% which could be the behaviour $1$ or some more complex behaviours. \\
Indeed, following the correspondance between designs and proofs of the hypersequentialized polarised linear logic $H$ which is given in the annex, the design ${\cal D}$ may be seen as an attempt to prove the formula $S=S_1\oplus S_2$ where $S_1$ and $S_2$ are the (proto) - formulas associated with the first and second reading of $S$ in their linear and hypersequentialized formulations : 
 
 \begin{center}
 \footnotesize{
\shortstack{
\shortstack{${\cal D}_{d'}$\\
$\vdots$}
\hspace{1em}
\shortstack{\shortstack{ $\downarrow A^\perp(e_d)\vdash$}\hspace{2em} 
\shortstack{ $\downarrow P^\perp(d,e_d)\vdash$}\\
$\hrulefill$\\
$ \vdash\downarrow L^\perp(d),\exists y ( \uparrow A(y)\otimes \uparrow P(d,y))$}
\hspace{1em}
\shortstack{${\cal D}_{d''}$\\
$\vdots$}\\
$\hrulefill$\\
$(\forall x  (\uparrow L(x)\linefle \exists y  ( \uparrow A(y)\otimes\uparrow P(x,y))))^\perp\vdash$\\
$\hrulefill$\\
$\vdash S$}
}
\end{center}
\noindent We retrieve the semantical notion of ``logical form'' but resting on \emph{behaviours} instead of, simply, logical formulae. There are finally two possible ways to associate a behaviour with a sentence:
\begin{itemize}
\item[-] either, we can consider that the design obtained (as in the previous section) as a minimal justification of $S$ may generate a behaviour associated with $S$. Thus ${\mathbb S}_{gen}={\cal D}^{\perp\perp}$.
 
\item[-]  or we can consider that the behaviour associated with $S$ corresponds to the linear formula (in an hypersequentialised formulation):
\begin{center}
 ${\mathbb S}=(\forall x (\downarrow {\mathbb L}(x)\linefle \exists y(\downarrow {\mathbb A}(y)\otimes \downarrow {\mathbb P}(x,y))))\oplus \exists y (\downarrow {\mathbb A}(y)\otimes\forall x \uparrow (\downarrow {\mathbb L}(x)\linefle \downarrow{\mathbb P}(x,y)))$.
\end{center}
\end{itemize}
The later interpretation of $S$'s meaning is in fact a \emph{family of behaviours} because ${\mathbb S}$ depends on the behaviours  $ {\mathbb L}(x)$ , ${\mathbb A}(y)$ and $ {\mathbb P}(x,y)))$. 

\rem 
As a \emph{logical formula}, $S$ is seen as the disjunction of $S_1$ and $S_2$ , namely as the formula $
 S=  S_1\oplus \downarrow S_2$, and as a \emph{behaviour}, seen as the union\footnote{This is one of the mains results of Ludics: the internal completeness ensures that the elementary operation of union is enough to obtain all the designs of the disjunction.} of the two behaviours associated with the two terms of the disjunct. Hence we get a logical account of the fact that interaction may activate only one of both logical sub-formulas, depending on the scope of ``some''.

\rem The behaviour ${\mathbb S}_{gen}$ contains all the behaviours logically built  from the behaviours associated with the elemantary pieces $L(x)$, $A(y)$ and $P(x,y)$. This way we get a first (and still rough) account of the logical particles of meaning.\\
%\rem Notice that ${\mathbb S}_{gen}$ contains all the element of the family obtained by logical decomposition of $S$.  
\rm

\noindent Finally, Ludics enables us to go further into the specification of the logical form. \\

\noindent{\bf Decomposing ``atomic formulas''}
 \begin{enumerate}
 \item It is of course possible to consider the leaves of a decomposition as atomic formulae, if decomposition ends up. In this case, they are seen as \emph{data items}\footnote{in Ludics this is possible by means of the use of the linear multiplicative constant $1$}.\\
  We can  thus consider the following design ${\cal D}'$ as a justification of $S$ :
  
   \begin{center}
 \footnotesize{
\shortstack{
\shortstack{${\cal D}_{d'}$\\
$\vdots$}
\hspace{1em}
\shortstack{\shortstack{  $\hrulefill_\emptyset$\\
$\vdash A(e_{d})$\\
$\hrulefill$\\
$\downarrow A^\perp(e_{d})\vdash$}\hspace{2em} 
\shortstack{ $\hrulefill_\emptyset$\\
$\vdash P(d,e_{d})$\\
$\hrulefill$\\
$\downarrow P^\perp(d,e_{d})\vdash$}\\
 $\hrulefill$\\
$ \vdash\downarrow L^\perp(d),\exists y ( \uparrow A(y)\otimes \uparrow P(d,y))$}
\hspace{1em}
\shortstack{${\cal D}_{d''}$\\
$\vdots$}\\
$\hrulefill$\\
$(\forall x  (\uparrow L(x)\linefle \exists y  ( \uparrow A(y)\otimes\uparrow P(x,y))))^\perp\vdash$}
}
\end{center}
  
Let us remark that ${\cal D}'$  is more defined than ${\cal D}$. In Ludics this means that ${\cal D}^\perp\subset{\cal D}'^{\perp}$ and this may be understood here that the justification is more informative, more precise.

 \item But we may also consider that $L(x)$, $A(y)$ and $P(x,y)$ are \emph{still decomposable}.  That amounts to recognize that  ${\mathbb S}_1$ contains other designs: all those which are more defined than ${\cal D}$. Designs \emph{more defined than} ${\cal D}$ are built on the same schema than ${\cal D}'$ but instead of ending on the the empty set, they continue on non empty ramifications, thus allowing the 
exploration of $A(e_f)$ or $P(f,e_f)$ which were alleged atomic formulae in the previous designs.
 \end{enumerate}

\noindent
The vericonditional interpretation is here retrieved as an indirect (and secondary) consequence of our "(para)\-proofs as meanings"\footnote{In the opposition of two processes of proof search, both cannot be "real" proofs, it is the reason why we call them paraproofs}  interpretation because now, ${\cal D}'$ is really a proof provided that $A(f)$ and $P(f,e_f)$ are either data items, that is the true linear formula {\bf 1} or are  provable when they are decomposable.

\subsection{How to go further ?}
 
\noindent{\bf Towards speech acts - and the use of ${\cal F}$ax} 
Instead of simple yes/no questions, where convergence occurs for "yes" and divergence for ``no'', we may take so called \emph{wh}-questions into consideration, for example ``which is the african language that John speaks ?''. In this case \emph{we expect  that the interaction has the answer as its by-product (or its side effect)}.\\
To reach this goal, let us associate with such a question (that we may see as a \emph{speech act}) a design in which \emph{there is a locus for storing the answer}. In our formulation of designs as $HS$-paraproofs, this question will be associated with a paraproof of the sequent $S\vdash A$ where $A$ is a formula equal to $\uparrow A_1\oplus\dots\oplus \uparrow A_n$ corresponding to the logical form of  '' is some african language'' (afar, peul, ewe, ewondo...).
\vspace{0,5em}
\noindent
A complex  design using ${\cal F}ax$ will be associated with the question 
 ''which is the african language that John speaks ?'' .
The result  of the interaction of this design with ${\cal D}'$ is~:
{\footnotesize
\shortstack{$\hrulefill_\emptyset$\\
$\vdash A_e$\\
$\hrulefill$\\
$\downarrow A_e^\perp\vdash$\\
$\hrulefill$\\
$\vdash A$}
}
 which can be read as `` $A_e$ is this african language''
(where $A_e$ is the african language that 
 John speaks (in ${\cal D}'$)).
 
 \vspace{0,5em}
\noindent
In our opinion, this suggests a way to perform in Ludics a  unified treatment of Logical Forms and Speech Acts. At the same time, this underlines the richness of the ludical framework to give an account of the interactions in language.

\section{Conclusion}
\noindent
In this paper, we tried to give an account of Ludics and of the new way it allows to specify Meaning in Language: not by considerations on truth conditions but by using the important concept of \emph{interaction}. To access the meaning of a sentence is mainly to know how to question, to answer to or to refute this sentence, and to know how to extend the discourse (or the dialogue) to which it belongs. In such a conception, the meaning of a sentence is a moment inside an entire process which coud be conceived as infinite (if for instance we admit that the interpretation or the argumentation process with regards to any statement is potentially infinite). Ludics gives a precise form to these views by means of the notions of \emph{normalization} and \emph{behaviour}.\\
Otherwise, the emphasis put on \emph{loci} has, as a valuable consequence, the fact that we may conceive several instances of the same \emph{sign} (a sentence, a word etc.) as having various meanings, according to the location it has in a discourse or a dialogue, thus giving suggestions for dealing with many rhetorical figures (and \emph{fallacies}). The infinite design ${\cal F}ax$ allows to delocate such meanings but its use is not mandatory. Moreover, the fact (not much developed in this extended abstract) that a design may be viewed either as a kind of proof (in a syntactic setting of the framework) or as a game (in a semantic setting of it) provides us with interesting insights on Pragmatics and Wittgensteinian language games. In a pragmatic theory of presupposition, for instance, \emph{presupposing} implies making an assertion where the hearer has no access to a previous step made by the speaker, if (s)he rejects this step, (s)he makes the process to diverge. Other ''games'' may be explored. Wittgenstein for instance quoted \emph{elicitation}, that is the way in which somebody may obtain an answer to a question. Every time, ${\cal F}ax$ is used to transfer a meaning from a location to another one (for instance from the discourse or the brain of the other speaker to the one of the eliciter). Those games may be envisaged without any kind of ''winning strategy''. In a speech act seen as a game, there is no win, simply the appropriate use of some designs in order to reach an objective (which may be a common one). Future works will be done in those directions.
%\end{document}
\section{Annexe A : A hypersequentialized version of the linear sequent calculus}
%\subsubsection{Forks}
We give here a short presentation of a hypersequentialized version of linear calculus, which enables one to manipule the designs as (para)proofs of a logical calculus.
\subsection{Formulas and sequents}
\noindent
By means of polarity, we may simplify the calculus by keeping \emph{only positive formulae}. Of course, there are still negative formulae... but they are simply put on the left-hand side after they have been changed into their negation. Moreover, in order to make paraproofs to look like sequences of alternate steps (like it is the case in ordinary games), we will make blocks of positive and of negative formulae in such a way that each one is introduced in only one step, thus necessarily using \emph{synthetic connectives}. Such connectives are still denoted $\oplus$ and $\otimes$ but are of various arities. We  will distinguish the case where both $\oplus$ and $\otimes$ are of arity $1$ and denote it $\downarrow$.

\begin{itemize}
\item[-] The only linear formulae which are considered in such a sequent calculus are built from the set $P$ of linear constants and propositionnal variables according to the following schema~:
\[
F=P|(F^\perp\otimes\dots\otimes F^\perp)\oplus\dots\oplus (F^\perp\otimes\dots\otimes F^\perp)|
\downarrow F^\perp
\]
\item The sequents are {\bf denoted} $\Gamma\vdash\Delta$ where $\Delta$ is a multiset of formulas and $\Gamma$ contains at most a formula.
\end{itemize}

\subsection{Rules}
\begin{itemize}
\item There are some axioms (logical and non logical axioms):

\begin{center}
\shortstack{$\overline{P\vdash P}$}
\hspace{2em}
\shortstack{$\overline{\vdash 1}$}
\hspace{2em}
\shortstack{$\overline{\vdash \downarrow T,\Delta}$}
\hspace{2em}
\shortstack{$\hrulefill_\dag$\\
$\vdash\Delta$}

\end{center}
where $P$ is a propositionnal variable ; $1$ and $T$ are the usual linear constants (respectively positive and negative).
\item The "logical"  rules  are the following ones :

{\bf Negative rule}\\
\begin{center}
\shortstack{
\shortstack{$\vdash A_{11},\dots,A_{1n_1},\Gamma$}\hspace{1em}\dots \hspace{1em}\shortstack{$\vdash A_{p1},\dots,A_{pn_p}, \Gamma$}\\
$\hrulefill$\\
 $(A_{11}\otimes \dots \otimes A_{1n_1})\oplus\dots\oplus(A_{p1}\otimes\dots\otimes A_{pn_p})\vdash\Gamma$}
\end{center}

{\bf Positive rule}\\
\begin{center}
\shortstack{
\shortstack{$ A_{i 1}\vdash\Gamma_1$}\hspace{1em}\dots \shortstack{$ A_{in_i}\vdash \Gamma_p$}\\
$\hrulefill$\\
 $\vdash(A_{11}\otimes \dots \otimes A_{1n_1})\oplus\dots\oplus(A_{p1}\otimes\dots\otimes A_{pn_p}), \Gamma$}
\end{center} 

where $\cup \Gamma_k\subset\Gamma$ and for $k,l\in\{1,\dots p\}$ the $\Gamma_k\cap\Gamma_l=\emptyset$.
\end{itemize}
\subsection{Remarks on Shifts}

\noindent
Using the shift is a way to break a block of a given polarity. 
Separate steps may be enforced by using the \emph{shift} operators $\dw$ and $\uparrow$ which  change the negative (resp. positive) polarity into the positive (resp. negative) one.  The rules introducing such shifted formulas are particular cases of the positive and the negative one:
$$
\displaylines{
\hfill
\prooftree
A^{\bot}\vdash \Gamma
\justifies
\vdash\dw A, \Gamma
\using
[+]
\endprooftree
\hfill
\prooftree
\vdash A^{\bot}, \Gamma
\justifies
\dw A\vdash \Gamma
\using
[-]
\endprooftree
\hfill
}
$$
where $A$ is a negative formula.\\
\noindent{\bf Example}  In a block like $A\tms B\tms C $ in principle, $A, B$ and $C$ are negative, but if we don't want to deal with $A, B, C$ simultaneously, we may change the polarity of $B\tms C$ (which is positive) and make it negative by means of $\uparrow$. We write then $A\tms \uparrow (B\tms C)$.\\ 
Compare the two following partial proofs, where (1) does not use any shifts and (2) uses one : 

instead of (1): 
\shortstack{$A^\perp\vdash$\hspace{1em} $B^\perp\vdash$\hspace{1em}$C^\perp\vdash$\\
$\hrulefill$\\
$\vdash A\otimes B\otimes C$}
we get (2) :
\shortstack{$A^\perp\vdash$\hspace{1em}\shortstack{$B^\perp\vdash$\hspace{1em}$C^\perp\vdash$\\
$\hrulefill$\\
$\vdash B\otimes C$\\
$\hrulefill$\\
$\downarrow(B\otimes C)^\perp\vdash$}\\
$\hrulefill$\\
$\vdash A\otimes\uparrow(B\otimes C)$}\\

\noindent
We may use the notation $\oplus_y$ (and dually $\&_x$)  
instead of $F_{y_1}\oplus\dots\oplus F_{y_n}$ or simply $\exists y$  
(dually $\forall x$) when it is clear in context that $y$ belongs to a  
finite set.

\end{document}

%% file: nprooftree.tex
\newdimen\proofrulebreadth \proofrulebreadth=.05em
\newdimen\proofdotseparation \proofdotseparation=1.25ex
\newdimen\proofrulebaseline \proofrulebaseline=2ex
\newcount\proofdotnumber \proofdotnumber=3
\let\then\relax
\def\hfi{\hskip0pt plus.0001fil}
\mathchardef\squigto="3A3B
\newif\ifinsideprooftree\insideprooftreefalse
\newif\ifonleftofproofrule\onleftofproofrulefalse
\newif\ifproofdots\proofdotsfalse
\newif\ifdoubleproof\doubleprooffalse
\let\wereinproofbit\relax
\newdimen\shortenproofleft
\newdimen\shortenproofright
\newdimen\proofbelowshift
\newbox\proofabove
\newbox\proofbelow
\newbox\proofrulename
\def\shiftproofbelow{\let\next\relax\afterassignment\setshiftproofbelow\dimen0 }
\def\shiftproofbelowneg{\def\next{\multiply\dimen0 by-1 }%
\afterassignment\setshiftproofbelow\dimen0 }
\def\setshiftproofbelow{\next\proofbelowshift=\dimen0 }
\def\setproofrulebreadth{\proofrulebreadth}

\def\prooftree{% NESTED ZERO (\ifonleftofproofrule)
\ifnum	\lastpenalty=1
\then	\unpenalty
\else	\onleftofproofrulefalse
\fi
\ifonleftofproofrule
\else	\ifinsideprooftree
	\then	\hskip.5em plus1fil
	\fi
\fi
\bgroup% NESTED ONE (\proofbelow, \proofrulename, \proofabove,
\setbox\proofbelow=\hbox{}\setbox\proofrulename=\hbox{}%
\let\justifies\proofover\let\leadsto\proofoverdots\let\Justifies\proofoverdbl
\let\using\proofusing\let\[\prooftree
\ifinsideprooftree\let\]\endprooftree\fi
\proofdotsfalse\doubleprooffalse
\let\thickness\setproofrulebreadth
\let\shiftright\shiftproofbelow \let\shift\shiftproofbelow
\let\shiftleft\shiftproofbelowneg
\let\ifwasinsideprooftree\ifinsideprooftree
\insideprooftreetrue
\setbox\proofabove=\hbox\bgroup$\displaystyle % NESTED TWO
\let\wereinproofbit\prooftree
\shortenproofleft=0pt \shortenproofright=0pt \proofbelowshift=0pt
\onleftofproofruletrue\penalty1
}

\def\eproofbit{% NESTED TWO
\ifx	\wereinproofbit\prooftree
\then	\ifcase	\lastpenalty
	\then	\shortenproofright=0pt	% 0: some other object, no indentation
	\or	\unpenalty\hfil		% 1: empty hypotheses, just glue
	\or	\unpenalty\unskip	% 2: just had a tree, remove glue
	\else	\shortenproofright=0pt	% eh?
	\fi
\fi
\global\dimen0=\shortenproofleft
\global\dimen1=\shortenproofright
\global\dimen2=\proofrulebreadth
\global\dimen3=\proofbelowshift
\global\dimen4=\proofdotseparation
\global\count10=\proofdotnumber
$\egroup  % NESTED ONE
\shortenproofleft=\dimen0
\shortenproofright=\dimen1
\proofrulebreadth=\dimen2
\proofbelowshift=\dimen3
\proofdotseparation=\dimen4
\proofdotnumber=\count10
}

\def\proofover{% NESTED TWO
\eproofbit % NESTED ONE
\setbox\proofbelow=\hbox\bgroup % NESTED TWO
\let\wereinproofbit\proofover
$\displaystyle
}%
\def\proofoverdbl{% NESTED TWO
\eproofbit % NESTED ONE
\doubleprooftrue
\setbox\proofbelow=\hbox\bgroup % NESTED TWO
\let\wereinproofbit\proofoverdbl
$\displaystyle
}%
\def\proofoverdots{% NESTED TWO
\eproofbit % NESTED ONE
\proofdotstrue
\setbox\proofbelow=\hbox\bgroup % NESTED TWO
\let\wereinproofbit\proofoverdots
$\displaystyle
}%
\def\proofusing{% NESTED TWO
\eproofbit % NESTED ONE
\setbox\proofrulename=\hbox\bgroup % NESTED TWO
\let\wereinproofbit\proofusing
\kern0.3em$
}

\def\endprooftree{% NESTED TWO
\eproofbit % NESTED ONE
  \dimen5 =0pt% spread of hypotheses
\dimen0=\wd\proofabove \advance\dimen0-\shortenproofleft
\advance\dimen0-\shortenproofright
\dimen1=.5\dimen0 \advance\dimen1-.5\wd\proofbelow
\dimen4=\dimen1
\advance\dimen1\proofbelowshift \advance\dimen4-\proofbelowshift
\ifdim	\dimen1<0pt
\then	\advance\shortenproofleft\dimen1
	\advance\dimen0-\dimen1
	\dimen1=0pt
	\ifdim  \shortenproofleft<0pt
        \then   \setbox\proofabove=\hbox{%
			\kern-\shortenproofleft\unhbox\proofabove}%
                \shortenproofleft=0pt
        \fi
\fi
\ifdim	\dimen4<0pt
\then	\advance\shortenproofright\dimen4
	\advance\dimen0-\dimen4
	\dimen4=0pt
\fi
\ifdim	\shortenproofright<\wd\proofrulename
\then	\shortenproofright=\wd\proofrulename
\fi
\dimen2=\shortenproofleft \advance\dimen2 by\dimen1
\dimen3=\shortenproofright\advance\dimen3 by\dimen4
\ifproofdots
\then
	\dimen6=\shortenproofleft \advance\dimen6 .5\dimen0
	\setbox1=\vbox to\proofdotseparation{\vss\hbox{$\cdot$}\vss}%
	\setbox0=\hbox{%
		\advance\dimen6-.5\wd1
		\kern\dimen6
		$\vcenter to\proofdotnumber\proofdotseparation
			{\leaders\box1\vfill}$%
		\unhbox\proofrulename}%
\else	\dimen6=\fontdimen22\the\textfont2 % height of maths axis
	\dimen7=\dimen6
	\advance\dimen6by.5\proofrulebreadth
	\advance\dimen7by-.5\proofrulebreadth
	\setbox0=\hbox{%
		\kern\shortenproofleft
		\ifdoubleproof
		\then	\hbox to\dimen0{%
			$\mathsurround0pt\mathord=\mkern-6mu%
			\cleaders\hbox{$\mkern-2mu=\mkern-2mu$}\hfill
			\mkern-6mu\mathord=$}%
		\else	\vrule height\dimen6 depth-\dimen7 width\dimen0
		\fi
		\unhbox\proofrulename}%
	\ht0=\dimen6 \dp0=-\dimen7
\fi
\let\doll\relax
\ifwasinsideprooftree
\then	\let\VBOX\vbox
\else	\ifmmode\else$\let\doll=$\fi
	\let\VBOX\vcenter
\fi
\VBOX	{\baselineskip\proofrulebaseline \lineskip.2ex
	\expandafter\lineskiplimit\ifproofdots0ex\else-0.6ex\fi
	\hbox	spread\dimen5	{\hfi\unhbox\proofabove\hfi}%
	\hbox{\box0}%
	\hbox	{\kern\dimen2 \box\proofbelow}}\doll%
\global\dimen2=\dimen2
\global\dimen3=\dimen3
\egroup % NESTED ZERO
\ifonleftofproofrule
\then	\shortenproofleft=\dimen2
\fi
\shortenproofright=\dimen3
\onleftofproofrulefalse
\ifinsideprooftree
\then	\hskip.5em plus 1fil \penalty2
\fi
}

%% file: qobitree.tex
%% Copyright 1996 Jeffrey Mark Siskind
%
% This work may be distributed and/or modified under the
% conditions of the LaTeX Project Public License, either version 1.3
% of this license or (at your option) any later version.
% The latest version of this license is in
%   http://www.latex-project.org/lppl.txt
% and version 1.3 or later is part of all distributions of LaTeX
% version 2005/12/01 or later.
%
% This work has the LPPL maintenance status `unmaintained'.
%
% This work consists of all files listed in manifest.txt.
%
%% Copyright and licensing notice added 2008/11/29 by
%% Clea F. Rees on behalf of Jeffrey Mark Siskind.
%
% QobiTree tree macros written by Jeffrey Mark Siskind (Qobi@CIS.UPenn.EDU)
%
\newcounter{treecount}
\newcounter{branchcount}
\setcounter{treecount}{0}
\newsavebox{\parentbox}
\newsavebox{\treebox}
\newsavebox{\treeboxone}
\newsavebox{\treeboxtwo}
\newsavebox{\treeboxthree}
\newsavebox{\treeboxfour}
\newsavebox{\treeboxfive}
\newsavebox{\treeboxsix}
\newsavebox{\treeboxseven}
\newsavebox{\treeboxeight}
\newsavebox{\treeboxnine}
\newsavebox{\treeboxten}
\newsavebox{\treeboxeleven}
\newsavebox{\treeboxtwelve}
\newsavebox{\treeboxthirteen}
\newsavebox{\treeboxfourteen}
\newsavebox{\treeboxfifteen}
\newsavebox{\treeboxsixteen}
\newsavebox{\treeboxseventeen}
\newsavebox{\treeboxeighteen}
\newsavebox{\treeboxnineteen}
\newsavebox{\treeboxtwenty}
\newlength{\treeoffsetone}
\newlength{\treeoffsettwo}
\newlength{\treeoffsetthree}
\newlength{\treeoffsetfour}
\newlength{\treeoffsetfive}
\newlength{\treeoffsetsix}
\newlength{\treeoffsetseven}
\newlength{\treeoffseteight}
\newlength{\treeoffsetnine}
\newlength{\treeoffsetten}
\newlength{\treeoffseteleven}
\newlength{\treeoffsettwelve}
\newlength{\treeoffsetthirteen}
\newlength{\treeoffsetfourteen}
\newlength{\treeoffsetfifteen}
\newlength{\treeoffsetsixteen}
\newlength{\treeoffsetseventeen}
\newlength{\treeoffseteighteen}
\newlength{\treeoffsetnineteen}
\newlength{\treeoffsettwenty}

\newlength{\treeshiftone}
\newlength{\treeshifttwo}
\newlength{\treeshiftthree}
\newlength{\treeshiftfour}
\newlength{\treeshiftfive}
\newlength{\treeshiftsix}
\newlength{\treeshiftseven}
\newlength{\treeshifteight}
\newlength{\treeshiftnine}
\newlength{\treeshiftten}
\newlength{\treeshifteleven}
\newlength{\treeshifttwelve}
\newlength{\treeshiftthirteen}
\newlength{\treeshiftfourteen}
\newlength{\treeshiftfifteen}
\newlength{\treeshiftsixteen}
\newlength{\treeshiftseventeen}
\newlength{\treeshifteighteen}
\newlength{\treeshiftnineteen}
\newlength{\treeshifttwenty}
\newlength{\treewidthone}
\newlength{\treewidthtwo}
\newlength{\treewidththree}
\newlength{\treewidthfour}
\newlength{\treewidthfive}
\newlength{\treewidthsix}
\newlength{\treewidthseven}
\newlength{\treewidtheight}
\newlength{\treewidthnine}
\newlength{\treewidthten}
\newlength{\treewidtheleven}
\newlength{\treewidthtwelve}
\newlength{\treewidththirteen}
\newlength{\treewidthfourteen}
\newlength{\treewidthfifteen}
\newlength{\treewidthsixteen}
\newlength{\treewidthseventeen}
\newlength{\treewidtheighteen}
\newlength{\treewidthnineteen}
\newlength{\treewidthtwenty}
\newlength{\daughteroffsetone}
\newlength{\daughteroffsettwo}
\newlength{\daughteroffsetthree}
\newlength{\daughteroffsetfour}
\newlength{\branchwidthone}
\newlength{\branchwidthtwo}
\newlength{\branchwidththree}
\newlength{\branchwidthfour}
\newlength{\parentoffset}
\newlength{\treeoffset}
\newlength{\daughteroffset}
\newlength{\branchwidth}
\newlength{\parentwidth}
\newlength{\treewidth}
\newcommand{\ontop}[1]{\begin{tabular}{c}#1\end{tabular}}
\newcommand{\poptree}{%
\ifnum\value{treecount}=0\typeout{QobiTeX warning---Tree stack underflow}\fi%
\addtocounter{treecount}{-1}%
\setlength{\treeoffsettwo}{\treeoffsetthree}%
\setlength{\treeoffsetthree}{\treeoffsetfour}%
\setlength{\treeoffsetfour}{\treeoffsetfive}%
\setlength{\treeoffsetfive}{\treeoffsetsix}%
\setlength{\treeoffsetsix}{\treeoffsetseven}%
\setlength{\treeoffsetseven}{\treeoffseteight}%
\setlength{\treeoffseteight}{\treeoffsetnine}%
\setlength{\treeoffsetnine}{\treeoffsetten}%
\setlength{\treeoffsetten}{\treeoffseteleven}%
\setlength{\treeoffseteleven}{\treeoffsettwelve}%
\setlength{\treeoffsettwelve}{\treeoffsetthirteen}%
\setlength{\treeoffsetthirteen}{\treeoffsetfourteen}%
\setlength{\treeoffsetfourteen}{\treeoffsetfifteen}%
\setlength{\treeoffsetfifteen}{\treeoffsetsixteen}%
\setlength{\treeoffsetsixteen}{\treeoffsetseventeen}%
\setlength{\treeoffsetseventeen}{\treeoffseteighteen}%
\setlength{\treeoffseteighteen}{\treeoffsetnineteen}%
\setlength{\treeoffsetnineteen}{\treeoffsettwenty}%
\setlength{\treeshifttwo}{\treeshiftthree}%
\setlength{\treeshiftthree}{\treeshiftfour}%
\setlength{\treeshiftfour}{\treeshiftfive}%
\setlength{\treeshiftfive}{\treeshiftsix}%
\setlength{\treeshiftsix}{\treeshiftseven}%
\setlength{\treeshiftseven}{\treeshifteight}%
\setlength{\treeshifteight}{\treeshiftnine}%
\setlength{\treeshiftnine}{\treeshiftten}%
\setlength{\treeshiftten}{\treeshifteleven}%
\setlength{\treeshifteleven}{\treeshifttwelve}%
\setlength{\treeshifttwelve}{\treeshiftthirteen}%
\setlength{\treeshiftthirteen}{\treeshiftfourteen}%
\setlength{\treeshiftfourteen}{\treeshiftfifteen}%
\setlength{\treeshiftfifteen}{\treeshiftsixteen}%
\setlength{\treeshiftsixteen}{\treeshiftseventeen}%
\setlength{\treeshiftseventeen}{\treeshifteighteen}%
\setlength{\treeshifteighteen}{\treeshiftnineteen}%
\setlength{\treeshiftnineteen}{\treeshifttwenty}%
\setlength{\treewidthtwo}{\treewidththree}%
\setlength{\treewidththree}{\treewidthfour}%
\setlength{\treewidthfour}{\treewidthfive}%
\setlength{\treewidthfive}{\treewidthsix}%
\setlength{\treewidthsix}{\treewidthseven}%
\setlength{\treewidthseven}{\treewidtheight}%
\setlength{\treewidtheight}{\treewidthnine}%
\setlength{\treewidthnine}{\treewidthten}%
\setlength{\treewidthten}{\treewidtheleven}%
\setlength{\treewidtheleven}{\treewidthtwelve}%
\setlength{\treewidthtwelve}{\treewidththirteen}%
\setlength{\treewidththirteen}{\treewidthfourteen}%
\setlength{\treewidthfourteen}{\treewidthfifteen}%
\setlength{\treewidthfifteen}{\treewidthsixteen}%
\setlength{\treewidthsixteen}{\treewidthseventeen}%
\setlength{\treewidthseventeen}{\treewidtheighteen}%
\setlength{\treewidtheighteen}{\treewidthnineteen}%
\setlength{\treewidthnineteen}{\treewidthtwenty}%
\sbox{\treeboxtwo}{\usebox{\treeboxthree}}%
\sbox{\treeboxthree}{\usebox{\treeboxfour}}%
\sbox{\treeboxfour}{\usebox{\treeboxfive}}%
\sbox{\treeboxfive}{\usebox{\treeboxsix}}%
\sbox{\treeboxsix}{\usebox{\treeboxseven}}%
\sbox{\treeboxseven}{\usebox{\treeboxeight}}%
\sbox{\treeboxeight}{\usebox{\treeboxnine}}%
\sbox{\treeboxnine}{\usebox{\treeboxten}}%
\sbox{\treeboxten}{\usebox{\treeboxeleven}}%
\sbox{\treeboxeleven}{\usebox{\treeboxtwelve}}%
\sbox{\treeboxtwelve}{\usebox{\treeboxthirteen}}%
\sbox{\treeboxthirteen}{\usebox{\treeboxfourteen}}%
\sbox{\treeboxfourteen}{\usebox{\treeboxfifteen}}%
\sbox{\treeboxfifteen}{\usebox{\treeboxsixteen}}%
\sbox{\treeboxsixteen}{\usebox{\treeboxseventeen}}%
\sbox{\treeboxseventeen}{\usebox{\treeboxeighteen}}%
\sbox{\treeboxeighteen}{\usebox{\treeboxnineteen}}%
\sbox{\treeboxnineteen}{\usebox{\treeboxtwenty}}}
\newcommand{\leaf}[1]{%
\ifnum\value{treecount}=20\typeout{QobiTeX warning---Tree stack overflow}\fi%
\addtocounter{treecount}{1}%
\sbox{\treeboxtwenty}{\usebox{\treeboxnineteen}}%
\sbox{\treeboxnineteen}{\usebox{\treeboxeighteen}}%
\sbox{\treeboxeighteen}{\usebox{\treeboxseventeen}}%
\sbox{\treeboxseventeen}{\usebox{\treeboxsixteen}}%
\sbox{\treeboxsixteen}{\usebox{\treeboxfifteen}}%
\sbox{\treeboxfifteen}{\usebox{\treeboxfourteen}}%
\sbox{\treeboxfourteen}{\usebox{\treeboxthirteen}}%
\sbox{\treeboxthirteen}{\usebox{\treeboxtwelve}}%
\sbox{\treeboxtwelve}{\usebox{\treeboxeleven}}%
\sbox{\treeboxeleven}{\usebox{\treeboxten}}%
\sbox{\treeboxten}{\usebox{\treeboxnine}}%
\sbox{\treeboxnine}{\usebox{\treeboxeight}}%
\sbox{\treeboxeight}{\usebox{\treeboxseven}}%
\sbox{\treeboxseven}{\usebox{\treeboxsix}}%
\sbox{\treeboxsix}{\usebox{\treeboxfive}}%
\sbox{\treeboxfive}{\usebox{\treeboxfour}}%
\sbox{\treeboxfour}{\usebox{\treeboxthree}}%
\sbox{\treeboxthree}{\usebox{\treeboxtwo}}%
\sbox{\treeboxtwo}{\usebox{\treeboxone}}%
\sbox{\treeboxone}{\ontop{#1}}%
\sbox{\treeboxone}{\raisebox{-\ht\treeboxone}{\usebox{\treeboxone}}}%
\setlength{\treeoffsettwenty}{\treeoffsetnineteen}%
\setlength{\treeoffsetnineteen}{\treeoffseteighteen}%
\setlength{\treeoffseteighteen}{\treeoffsetseventeen}%
\setlength{\treeoffsetseventeen}{\treeoffsetsixteen}%
\setlength{\treeoffsetsixteen}{\treeoffsetfifteen}%
\setlength{\treeoffsetfifteen}{\treeoffsetfourteen}%
\setlength{\treeoffsetfourteen}{\treeoffsetthirteen}%
\setlength{\treeoffsetthirteen}{\treeoffsettwelve}%
\setlength{\treeoffsettwelve}{\treeoffseteleven}%
\setlength{\treeoffseteleven}{\treeoffsetten}%
\setlength{\treeoffsetten}{\treeoffsetnine}%
\setlength{\treeoffsetnine}{\treeoffseteight}%
\setlength{\treeoffseteight}{\treeoffsetseven}%
\setlength{\treeoffsetseven}{\treeoffsetsix}%
\setlength{\treeoffsetsix}{\treeoffsetfive}%
\setlength{\treeoffsetfive}{\treeoffsetfour}%
\setlength{\treeoffsetfour}{\treeoffsetthree}%
\setlength{\treeoffsetthree}{\treeoffsettwo}%
\setlength{\treeoffsettwo}{\treeoffsetone}%
\setlength{\treeoffsetone}{0.5\wd\treeboxone}%
\setlength{\treeshifttwenty}{\treeshiftnineteen}%
\setlength{\treeshiftnineteen}{\treeshifteighteen}%
\setlength{\treeshifteighteen}{\treeshiftseventeen}%
\setlength{\treeshiftseventeen}{\treeshiftsixteen}%
\setlength{\treeshiftsixteen}{\treeshiftfifteen}%
\setlength{\treeshiftfifteen}{\treeshiftfourteen}%
\setlength{\treeshiftfourteen}{\treeshiftthirteen}%
\setlength{\treeshiftthirteen}{\treeshifttwelve}%
\setlength{\treeshifttwelve}{\treeshifteleven}%
\setlength{\treeshifteleven}{\treeshiftten}%
\setlength{\treeshiftten}{\treeshiftnine}%
\setlength{\treeshiftnine}{\treeshifteight}%
\setlength{\treeshifteight}{\treeshiftseven}%
\setlength{\treeshiftseven}{\treeshiftsix}%
\setlength{\treeshiftsix}{\treeshiftfive}%
\setlength{\treeshiftfive}{\treeshiftfour}%
\setlength{\treeshiftfour}{\treeshiftthree}%
\setlength{\treeshiftthree}{\treeshifttwo}%
\setlength{\treeshifttwo}{\treeshiftone}%
\setlength{\treeshiftone}{0pt}%
\setlength{\treewidthtwenty}{\treewidthnineteen}%
\setlength{\treewidthnineteen}{\treewidtheighteen}%
\setlength{\treewidtheighteen}{\treewidthseventeen}%
\setlength{\treewidthseventeen}{\treewidthsixteen}%
\setlength{\treewidthsixteen}{\treewidthfifteen}%
\setlength{\treewidthfifteen}{\treewidthfourteen}%
\setlength{\treewidthfourteen}{\treewidththirteen}%
\setlength{\treewidththirteen}{\treewidthtwelve}%
\setlength{\treewidthtwelve}{\treewidtheleven}%
\setlength{\treewidtheleven}{\treewidthten}%
\setlength{\treewidthten}{\treewidthnine}%
\setlength{\treewidthnine}{\treewidtheight}%
\setlength{\treewidtheight}{\treewidthseven}%
\setlength{\treewidthseven}{\treewidthsix}%
\setlength{\treewidthsix}{\treewidthfive}%
\setlength{\treewidthfive}{\treewidthfour}%
\setlength{\treewidthfour}{\treewidththree}%
\setlength{\treewidththree}{\treewidthtwo}%
\setlength{\treewidthtwo}{\treewidthone}%
\setlength{\treewidthone}{\wd\treeboxone}}
\newcommand{\branch}[2]{%
\setcounter{branchcount}{#1}%
\ifnum\value{branchcount}=1\sbox{\parentbox}{\ontop{#2}}%
\setlength{\parentoffset}{\treeoffsetone}%
\addtolength{\parentoffset}{-0.5\wd\parentbox}%
\setlength{\daughteroffset}{0in}%
\ifdim\parentoffset<0in%
\setlength{\daughteroffset}{-\parentoffset}%
\setlength{\parentoffset}{0in}\fi%
\setlength{\parentwidth}{\parentoffset}%
\addtolength{\parentwidth}{\wd\parentbox}%
\setlength{\treeoffset}{\daughteroffset}%
\addtolength{\treeoffset}{\treeoffsetone}%
\setlength{\treewidth}{\wd\treeboxone}%
\addtolength{\treewidth}{\daughteroffset}%
\ifdim\treewidth<\parentwidth\setlength{\treewidth}{\parentwidth}\fi%
\sbox{\treebox}{\begin{minipage}{\treewidth}%
\begin{flushleft}%
\hspace*{\parentoffset}\usebox{\parentbox}\\
{\setlength{\unitlength}{2ex}%
\hspace*{\treeoffset}\begin{picture}(0,1)%
\put(0,0){\line(0,1){1}}%
\end{picture}}\\
\vspace{-\baselineskip}
\hspace*{\daughteroffset}%
\raisebox{-\ht\treeboxone}{\usebox{\treeboxone}}%
\end{flushleft}%
\end{minipage}}%
\setlength{\treeoffsetone}{\parentoffset}%
\addtolength{\treeoffsetone}{0.5\wd\parentbox}%
\setlength{\treeshiftone}{0pt}%
\setlength{\treewidthone}{\treewidth}%
\sbox{\treeboxone}{\usebox{\treebox}}%
\else\ifnum\value{branchcount}=2\sbox{\parentbox}{\ontop{#2}}%
\setlength{\branchwidthone}{\treewidthtwo}%
\addtolength{\branchwidthone}{\treeoffsetone}%
\addtolength{\branchwidthone}{-\treeshiftone}%
\addtolength{\branchwidthone}{-\treeoffsettwo}%
\setlength{\branchwidth}{\branchwidthone}%
\setlength{\daughteroffsetone}{\branchwidth}%
\addtolength{\daughteroffsetone}{-\branchwidthone}%
\addtolength{\daughteroffsetone}{-\treeshiftone}%
\setlength{\parentoffset}{-0.5\wd\parentbox}%
\addtolength{\parentoffset}{\treeoffsettwo}%
\addtolength{\parentoffset}{0.5\branchwidth}%
\setlength{\daughteroffset}{0in}%
\ifdim\parentoffset<0in%
\setlength{\daughteroffset}{-\parentoffset}%
\setlength{\parentoffset}{0in}\fi%
\setlength{\parentwidth}{\parentoffset}%
\addtolength{\parentwidth}{\wd\parentbox}%
\setlength{\treeoffset}{\daughteroffset}%
\addtolength{\treeoffset}{\treeoffsettwo}%
\setlength{\treewidth}{\wd\treeboxone}%
\addtolength{\treewidth}{\daughteroffsetone}%
\addtolength{\treewidth}{\treewidthtwo}%
\addtolength{\treewidth}{\daughteroffset}%
\ifdim\treewidth<\parentwidth\setlength{\treewidth}{\parentwidth}\fi%
\sbox{\treebox}{\begin{minipage}{\treewidth}%
\begin{flushleft}%
\hspace*{\parentoffset}\usebox{\parentbox}\\
{\setlength{\unitlength}{0.5\branchwidth}%
\hspace*{\treeoffset}\begin{picture}(2,0.5)%
\put(0,0){\line(2,1){1}}%
\put(2,0){\line(-2,1){1}}%
\end{picture}}\\
\vspace{-\baselineskip}
\hspace*{\daughteroffset}%
\makebox[\treewidthtwo][l]%
{\raisebox{-\ht\treeboxtwo}{\usebox{\treeboxtwo}}}%
\hspace*{\daughteroffsetone}%
\raisebox{-\ht\treeboxone}{\usebox{\treeboxone}}%
\end{flushleft}%
\end{minipage}}%
\setlength{\treeoffsetone}{\parentoffset}%
\addtolength{\treeoffsetone}{0.5\wd\parentbox}%
\setlength{\treeshiftone}{0pt}%
\setlength{\treewidthone}{\treewidth}%
\sbox{\treeboxone}{\usebox{\treebox}}\poptree%
\else\ifnum\value{branchcount}=3\sbox{\parentbox}{\ontop{#2}}%
\setlength{\branchwidthone}{\treewidthtwo}%
\addtolength{\branchwidthone}{\treeoffsetone}%
\addtolength{\branchwidthone}{-\treeshiftone}%
\addtolength{\branchwidthone}{-\treeoffsettwo}%
\setlength{\branchwidthtwo}{\treewidththree}%
\addtolength{\branchwidthtwo}{\treeoffsettwo}%
\addtolength{\branchwidthtwo}{-\treeshifttwo}%
\addtolength{\branchwidthtwo}{-\treeoffsetthree}%
\setlength{\branchwidth}{\branchwidthone}%
\ifdim\branchwidthtwo>\branchwidth%
\setlength{\branchwidth}{\branchwidthtwo}\fi%
\setlength{\daughteroffsetone}{\branchwidth}%
\addtolength{\daughteroffsetone}{-\branchwidthone}%
\addtolength{\daughteroffsetone}{-\treeshiftone}%
\setlength{\daughteroffsettwo}{\branchwidth}%
\addtolength{\daughteroffsettwo}{-\branchwidthtwo}%
\addtolength{\daughteroffsettwo}{-\treeshifttwo}%
\setlength{\parentoffset}{-0.5\wd\parentbox}%
\addtolength{\parentoffset}{\treeoffsetthree}%
\addtolength{\parentoffset}{\branchwidth}%
\setlength{\daughteroffset}{0in}%
\ifdim\parentoffset<0in%
\setlength{\daughteroffset}{-\parentoffset}%
\setlength{\parentoffset}{0in}\fi%
\setlength{\parentwidth}{\parentoffset}%
\addtolength{\parentwidth}{\wd\parentbox}%
\setlength{\treeoffset}{\daughteroffset}%
\addtolength{\treeoffset}{\treeoffsetthree}%
\setlength{\treewidth}{\wd\treeboxone}%
\addtolength{\treewidth}{\daughteroffsetone}%
\addtolength{\treewidth}{\treewidthtwo}%
\addtolength{\treewidth}{\daughteroffsettwo}%
\addtolength{\treewidth}{\treewidththree}%
\addtolength{\treewidth}{\daughteroffset}%
\ifdim\treewidth<\parentwidth\setlength{\treewidth}{\parentwidth}\fi%
\sbox{\treebox}{\begin{minipage}{\treewidth}%
\begin{flushleft}%
\hspace*{\parentoffset}\usebox{\parentbox}\\
{\setlength{\unitlength}{0.5\branchwidth}%
\hspace*{\treeoffset}\begin{picture}(4,1)%
\put(0,0){\line(2,1){2}}%
\put(2,0){\line(0,1){1}}%
\put(4,0){\line(-2,1){2}}%
\end{picture}}\\
\vspace{-\baselineskip}
\hspace*{\daughteroffset}%
\makebox[\treewidththree][l]%
{\raisebox{-\ht\treeboxthree}{\usebox{\treeboxthree}}}%
\hspace*{\daughteroffsettwo}%
\makebox[\treewidthtwo][l]%
{\raisebox{-\ht\treeboxtwo}{\usebox{\treeboxtwo}}}%
\hspace*{\daughteroffsetone}%
\raisebox{-\ht\treeboxone}{\usebox{\treeboxone}}%
\end{flushleft}%
\end{minipage}}%
\setlength{\treeoffsetone}{\parentoffset}%
\addtolength{\treeoffsetone}{0.5\wd\parentbox}%
\setlength{\treeshiftone}{0pt}%
\setlength{\treewidthone}{\treewidth}%
\sbox{\treeboxone}{\usebox{\treebox}}\poptree\poptree%
\else\ifnum\value{branchcount}=4\sbox{\parentbox}{\ontop{#2}}%
\setlength{\branchwidthone}{\treewidthtwo}%
\addtolength{\branchwidthone}{\treeoffsetone}%
\addtolength{\branchwidthone}{-\treeshiftone}%
\addtolength{\branchwidthone}{-\treeoffsettwo}%
\setlength{\branchwidthtwo}{\treewidththree}%
\addtolength{\branchwidthtwo}{\treeoffsettwo}%
\addtolength{\branchwidthtwo}{-\treeshifttwo}%
\addtolength{\branchwidthtwo}{-\treeoffsetthree}%
\setlength{\branchwidththree}{\treewidthfour}%
\addtolength{\branchwidththree}{\treeoffsetthree}%
\addtolength{\branchwidththree}{-\treeshiftthree}%
\addtolength{\branchwidththree}{-\treeoffsetfour}%
\setlength{\branchwidth}{\branchwidthone}%
\ifdim\branchwidthtwo>\branchwidth%
\setlength{\branchwidth}{\branchwidthtwo}\fi%
\ifdim\branchwidththree>\branchwidth%
\setlength{\branchwidth}{\branchwidththree}\fi%
\setlength{\daughteroffsetone}{\branchwidth}%
\addtolength{\daughteroffsetone}{-\branchwidthone}%
\addtolength{\daughteroffsetone}{-\treeshiftone}%
\setlength{\daughteroffsettwo}{\branchwidth}%
\addtolength{\daughteroffsettwo}{-\branchwidthtwo}%
\addtolength{\daughteroffsettwo}{-\treeshifttwo}%
\setlength{\daughteroffsetthree}{\branchwidth}%
\addtolength{\daughteroffsetthree}{-\branchwidththree}%
\addtolength{\daughteroffsetthree}{-\treeshiftthree}%
\setlength{\parentoffset}{-0.5\wd\parentbox}%
\addtolength{\parentoffset}{\treeoffsetfour}%
\addtolength{\parentoffset}{1.5\branchwidth}%
\setlength{\daughteroffset}{0in}%
\ifdim\parentoffset<0in%
\setlength{\daughteroffset}{-\parentoffset}%
\setlength{\parentoffset}{0in}\fi%
\setlength{\parentwidth}{\parentoffset}%
\addtolength{\parentwidth}{\wd\parentbox}%
\setlength{\treeoffset}{\daughteroffset}%
\addtolength{\treeoffset}{\treeoffsetfour}%
\setlength{\treewidth}{\wd\treeboxone}%
\addtolength{\treewidth}{\daughteroffsetone}%
\addtolength{\treewidth}{\treewidthtwo}%
\addtolength{\treewidth}{\daughteroffsettwo}%
\addtolength{\treewidth}{\treewidththree}%
\addtolength{\treewidth}{\daughteroffsetthree}%
\addtolength{\treewidth}{\treewidthfour}%
\addtolength{\treewidth}{\daughteroffset}%
\ifdim\treewidth<\parentwidth\setlength{\treewidth}{\parentwidth}\fi%
\sbox{\treebox}{\begin{minipage}{\treewidth}%
\begin{flushleft}%
\hspace*{\parentoffset}\usebox{\parentbox}\\
{\setlength{\unitlength}{0.5\branchwidth}%
\hspace*{\treeoffset}\begin{picture}(6,1)%
\put(0,0){\line(3,1){3}}%
\put(2,0){\line(1,1){1}}%
\put(4,0){\line(-1,1){1}}%
\put(6,0){\line(-3,1){3}}%
\end{picture}}\\
\vspace{-\baselineskip}
\hspace*{\daughteroffset}%
\makebox[\treewidthfour][l]%
{\raisebox{-\ht\treeboxfour}{\usebox{\treeboxfour}}}%
\hspace*{\daughteroffsetthree}%
\makebox[\treewidththree][l]%
{\raisebox{-\ht\treeboxthree}{\usebox{\treeboxthree}}}%
\hspace*{\daughteroffsettwo}%
\makebox[\treewidthtwo][l]%
{\raisebox{-\ht\treeboxtwo}{\usebox{\treeboxtwo}}}%
\hspace*{\daughteroffsetone}%
\raisebox{-\ht\treeboxone}{\usebox{\treeboxone}}%
\end{flushleft}%
\end{minipage}}%
\setlength{\treeoffsetone}{\parentoffset}%
\addtolength{\treeoffsetone}{0.5\wd\parentbox}%
\setlength{\treeshiftone}{0pt}%
\setlength{\treewidthone}{\treewidth}%
\sbox{\treeboxone}{\usebox{\treebox}}\poptree\poptree\poptree%
\else\ifnum\value{branchcount}=5\sbox{\parentbox}{\ontop{#2}}%
\setlength{\branchwidthone}{\treewidthtwo}%
\addtolength{\branchwidthone}{\treeoffsetone}%
\addtolength{\branchwidthone}{-\treeshiftone}%
\addtolength{\branchwidthone}{-\treeoffsettwo}%
\setlength{\branchwidthtwo}{\treewidththree}%
\addtolength{\branchwidthtwo}{\treeoffsettwo}%
\addtolength{\branchwidthtwo}{-\treeshifttwo}%
\addtolength{\branchwidthtwo}{-\treeoffsetthree}%
\setlength{\branchwidththree}{\treewidthfour}%
\addtolength{\branchwidththree}{\treeoffsetthree}%
\addtolength{\branchwidththree}{-\treeshiftthree}%
\addtolength{\branchwidththree}{-\treeoffsetfour}%
\setlength{\branchwidthfour}{\treewidthfive}%
\addtolength{\branchwidthfour}{\treeoffsetfour}%
\addtolength{\branchwidthfour}{-\treeshiftfour}%
\addtolength{\branchwidthfour}{-\treeoffsetfive}%
\setlength{\branchwidth}{\branchwidthone}%
\ifdim\branchwidthtwo>\branchwidth%
\setlength{\branchwidth}{\branchwidthtwo}\fi%
\ifdim\branchwidththree>\branchwidth%
\setlength{\branchwidth}{\branchwidththree}\fi%
\ifdim\branchwidthfour>\branchwidth%
\setlength{\branchwidth}{\branchwidthfour}\fi%
\setlength{\daughteroffsetone}{\branchwidth}%
\addtolength{\daughteroffsetone}{-\branchwidthone}%
\addtolength{\daughteroffsetone}{-\treeshiftone}%
\setlength{\daughteroffsettwo}{\branchwidth}%
\addtolength{\daughteroffsettwo}{-\branchwidthtwo}%
\addtolength{\daughteroffsettwo}{-\treeshifttwo}%
\setlength{\daughteroffsetthree}{\branchwidth}%
\addtolength{\daughteroffsetthree}{-\branchwidththree}%
\addtolength{\daughteroffsetthree}{-\treeshiftthree}%
\setlength{\daughteroffsetfour}{\branchwidth}%
\addtolength{\daughteroffsetfour}{-\branchwidthfour}%
\addtolength{\daughteroffsetfour}{-\treeshiftfour}%
\setlength{\parentoffset}{-0.5\wd\parentbox}%
\addtolength{\parentoffset}{\treeoffsetfive}%
\addtolength{\parentoffset}{2\branchwidth}%
\setlength{\daughteroffset}{0in}%
\ifdim\parentoffset<0in%
\setlength{\daughteroffset}{-\parentoffset}%
\setlength{\parentoffset}{0in}\fi%
\setlength{\parentwidth}{\parentoffset}%
\addtolength{\parentwidth}{\wd\parentbox}%
\setlength{\treeoffset}{\daughteroffset}%
\addtolength{\treeoffset}{\treeoffsetfive}%
\setlength{\treewidth}{\wd\treeboxone}%
\addtolength{\treewidth}{\daughteroffsetone}%
\addtolength{\treewidth}{\treewidthtwo}%
\addtolength{\treewidth}{\daughteroffsettwo}%
\addtolength{\treewidth}{\treewidththree}%
\addtolength{\treewidth}{\daughteroffsetthree}%
\addtolength{\treewidth}{\treewidthfour}%
\addtolength{\treewidth}{\daughteroffsetfour}%
\addtolength{\treewidth}{\treewidthfive}%
\addtolength{\treewidth}{\daughteroffset}%
\ifdim\treewidth<\parentwidth\setlength{\treewidth}{\parentwidth}\fi%
\sbox{\treebox}{\begin{minipage}{\treewidth}%
\begin{flushleft}%
\hspace*{\parentoffset}\usebox{\parentbox}\\
{\setlength{\unitlength}{0.5\branchwidth}%
\hspace*{\treeoffset}\begin{picture}(8,1)%
\put(0,0){\line(4,1){4}}%
\put(2,0){\line(2,1){2}}%
\put(4,0){\line(0,1){1}}%
\put(6,0){\line(-2,1){2}}%
\put(8,0){\line(-4,1){4}}%
\end{picture}}\\
\vspace{-\baselineskip}
\hspace*{\daughteroffset}%
\makebox[\treewidthfive][l]%
{\raisebox{-\ht\treeboxfour}{\usebox{\treeboxfive}}}%
\hspace*{\daughteroffsetfour}%
\makebox[\treewidthfour][l]%
{\raisebox{-\ht\treeboxfour}{\usebox{\treeboxfour}}}%
\hspace*{\daughteroffsetthree}%
\makebox[\treewidththree][l]%
{\raisebox{-\ht\treeboxthree}{\usebox{\treeboxthree}}}%
\hspace*{\daughteroffsettwo}%
\makebox[\treewidthtwo][l]%
{\raisebox{-\ht\treeboxtwo}{\usebox{\treeboxtwo}}}%
\hspace*{\daughteroffsetone}%
\raisebox{-\ht\treeboxone}{\usebox{\treeboxone}}%
\end{flushleft}%
\end{minipage}}%
\setlength{\treeoffsetone}{\parentoffset}%
\addtolength{\treeoffsetone}{0.5\wd\parentbox}%
\setlength{\treeshiftone}{0pt}%
\setlength{\treewidthone}{\treewidth}%
\sbox{\treeboxone}{\usebox{\treebox}}\poptree\poptree\poptree\poptree%
\else\typeout{QobiTeX warning--- Can't handle #1 branching}\fi\fi\fi\fi\fi}
\newcommand{\faketreewidth}[1]{%
\sbox{\parentbox}{\ontop{#1}}%
\setlength{\treewidthone}{0.5\wd\parentbox}%
\addtolength{\treewidthone}{\treeoffsetone}%
\setlength{\treeshiftone}{\treeoffsetone}%
\addtolength{\treeshiftone}{-0.5\wd\parentbox}}
\newcommand{\tree}{%
\usebox{\treeboxone}
\setlength{\treeoffsetone}{\treeoffsettwo}%
\sbox{\treeboxone}{\usebox{\treeboxtwo}}%
\poptree}